\documentclass[preprint,3p,11pt]{elsarticle}

\usepackage{graphicx}
\usepackage{booktabs}
\usepackage{algorithm}
\usepackage{algorithmic}
\usepackage{multirow}
\usepackage{dblfloatfix}
\usepackage[table,dvipsnames]{xcolor}
\usepackage{hyperref}
\usepackage{amsmath}
\usepackage{amssymb}
\usepackage{fontawesome}
\usepackage{pdfpages}

\makeatletter
\def\ps@pprintTitle{%
  \let\@oddhead\@empty
  \let\@evenhead\@empty
  \let\@oddfoot\@empty
  \let\@evenfoot\@oddfoot
}
\makeatother

\begin{document}

\begin{frontmatter}
\title{Towards Desiderata-Driven Design of Visual Counterfactual Explainers}

\author[tub]{Sidney Bender}
\author[basf]{Jan Herrmann}
\author[bifold,tub,korea,mpi]{Klaus-Robert Müller}
\author[bifold,charite]{Grégoire Montavon\corref{cor1}}
\ead{gregoire.montavon@charite.de}

\cortext[cor1]{Corresponding author}

\address[tub]{Machine Learning Group, Technische Universit\"at Berlin, Germany}
\address[basf]{Statistics and Machine Learning, BASF SE, Ludwigshafen am Rhein, Germany}
\address[bifold]{BIFOLD\;--\;Berlin Institute for the Foundations of Learning and Data, Berlin, Germany}
\address[korea]{Department of Artificial Intelligence, Korea University, Seoul, Korea}
\address[mpi]{Max-Planck Institute for Informatics, Saarbruecken, Germany}
\address[charite]{Charit\'e\;--\;Universit\"atsmedizin Berlin, Germany}

\begin{abstract}
Visual counterfactual explainers (VCEs) are a straightforward and promising approach to enhancing the transparency of image classifiers. VCEs complement other types of explanations, such as feature attribution, by revealing the specific data transformations to which a machine learning model responds most strongly. In this paper, we argue that existing VCEs focus too narrowly on optimizing sample quality or change minimality; they fail to consider the more holistic desiderata for an explanation, such as fidelity, understandability, and sufficiency. To address this shortcoming, we explore new mechanisms for counterfactual generation and investigate how they can help fulfill these desiderata. We combine these mechanisms into a novel `smooth counterfactual explorer' (SCE) algorithm and demonstrate its effectiveness through systematic evaluations on synthetic and real data.
\end{abstract}

\end{frontmatter}

\section{Introduction}
The field of Explainable AI (XAI) \cite{DBLP:journals/aim/GunningA19,DBLP:journals/inffus/ArrietaRSBTBGGM20,DBLP:journals/pieee/SamekMLAM21} focuses on making ML models more transparent to their users.
It has given rise to a wide range of methods focusing on different data modalities and model types~(e.g.\ \cite{schnake2021higher,DBLP:journals/access/SahakyanAR21, Novakovsky2022, ali2022xai}), and can help to improve model performance~\cite{anders2022finding, weber2023beyond, linhardt2024preemptively}. XAI has also been  used in various application fields and scientific disciplines, e.g.\ medicine \cite{DBLP:conf/kdd/CaruanaLGKSE15,DBLP:journals/natmi/LundbergECDPNKH20,keyl2023single,klauschen-pathology24}, physics or chemistry \cite{DBLP:series/lncs/SchuttGTM19,keith2021combining,esders2025analyzing}
geoscience \cite{runge2019inferring,Toms2020,mamalakis2022investigating,hoffman2025evaluating} and history \cite{ElHajj2023,MartinezPandiani2023,eberle2024historical}, to unlock potential insight embodied by  machine learning models.

Counterfactuals explainers \cite{mothilal2020explaining, wachter2017counterfactual,dombrowski2022diffeomorphic,jeanneret2023adversarial} constitute a popular approach to explanation, which asks how input features would have to be changed to flip the predictor's decision. They have shown applicability in a broad range of settings ranging from linear models on tabular data (sometimes, this is also called algorithmic recourse) to sequence classification~\cite{naumann2021consequence}, graph classification~\cite{hansen2025graph}, unveiling deepfakes~\cite{yang2025unveiling}, object recognition~\cite{sun2023recursive}, counterfactual image generation~\cite{xia2024mitigating} or image regression~\cite{ha2025diffusion}. In this paper, we will focus on applications to image classification, and refer to the corresponding techniques as  `visual counterfactual explainers' (or VCEs) \cite{jeanneret2023adversarial,dombrowski2022diffeomorphic,rodriguez2021beyond}.

Explanations produced by VCEs take the form of an image (or a collection of images), which should differ from the original image only by the features that are necessary to effect a specific change at the output of the model. This can be e.g.\ the inclusion or removal of object parts, but also more intricate changes in image quality or color, that may not be accessible with other explanation techniques such as feature attribution. Another advantage of counterfactuals is that they are inherently actionable, e.g.\ together with a human in the loop, counterfactuals provide an implicit data augmentation scheme that can serve to address a model's missing invariances or reliance on spurious correlations~\cite{bender2023towards}. Mathematically, the search for counterfactuals can be formulated as an optimization problem:
\begin{align}
\arg\min_{\widetilde{x}} \quad d(x,\widetilde{x}) \qquad \text{s.t.} \quad f(\widetilde{x}) < 0 \quad \wedge \quad \widetilde{x} \in \mathcal{M},
\label{eq:counterfactual}
\end{align}
where $x$ is the data point (factual), $\widetilde{x}$ is the counterfactual, $\mathcal{M}$ is the data manifold, and $d$ is a distance along the manifold \cite{dombrowski2022diffeomorphic}. This effectively searches for the minimum perturbation of the original data point (according to some metric $d$) that flips the classification outcome.

\begin{figure}[t!]
    \centering
    \makebox[\textwidth][c]{\includegraphics[width=1.15\linewidth]{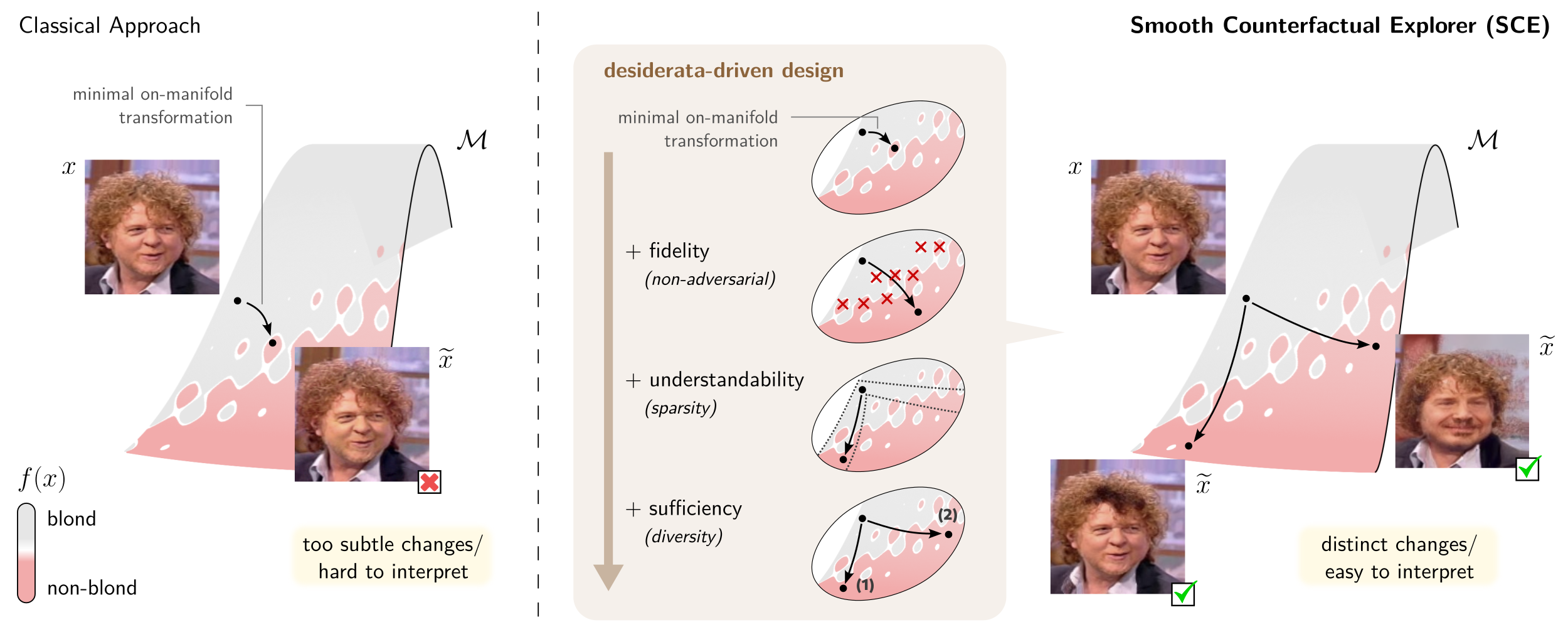}}
    \caption{Proposed desiderata-driven approach to counterfactual explanation, emphasizing `\textit{fidelity}' (the need to focus on the principal variations of $f$), `\textit{understandability}' (the need to align with the basis used for interpretation), and `\textit{sufficiency}' (the need to generate diverse counterfactuals). Compared to classical approaches based on minimizing distance on manifold, our desiderata-driven design of counterfactual explainers leads to more reliable and distinct explanations of what causes a change in class (in this example, the classifier changes its decision from `blond' to `non-blond' by darkening the hair or adding a beard).}
    \label{fig:intro}
\end{figure}

Despite these promising capabilities, we argue in this paper that current visual counterfactual explainers exhibit some major shortcomings that limit their potential use. Similar to Eq.\ \eqref{eq:counterfactual}'s formulation, they are mainly focused on optimizing two criteria: (1) staying on the data manifold $\mathcal{M}$, i.e.\ ensuring that $\widetilde{x}$ has high image {\em quality}, and (2) {\em minimality} of the transformation $x \mapsto \widetilde{x}$ through minimizing the distance function $d$. See Figure \ref{fig:intro} (left) for an illustration of a classical counterfactual search based on these criteria.

Higher image quality is arguably better, but one has to distinguish between \textit{image editing}~\cite{meng2021sdedit, couairon2022diffedit, wallace2023edict, huberman2024edit}, where the explicit goal is to produce high-quality edited images, and \textit{counterfactual explanations}, which are supposed to reveal the inner workings of a classifier.
Low-quality counterfactuals that clearly and correctly reveal the features used are arguably more valuable than high-quality counterfactuals that do not. For example, if the model is of `Clever Hans' type \cite{lapuschkin2019unmasking,geirhos2020shortcut,hermann2023foundations,kauffmann2025explainable} and spuriously relies e.g.\ on background features, the generated counterfactual must faithfully detect this flaw of the model and expose it in human-understandable manner.
Concerning the \textit{minimality} criterion, it is clear that a large distance between factual and counterfactual, e.g.\ resulting from choosing any data point on the other side of the decision boundary, would fail to disentangle relevant from irrelevant features. However, there are caveats to minimizing such distances. For example, suppose an image of a person is already on the boundary between the classes `smiling' and `serious'.
In that case, it is better for the sake of understandability to resolutely perturb one feature (e.g.\ the mouth) and leave the remaining features intact than to find a faint combination of unrelated pixels that produce a similar effect; in other words, sparsity in some meaningful semantic space should be preferred to minimality.

To increase the usefulness and actionability of counterfactual explanations, we propose to start from a well-established explanation desiderata by Swartout and Moore \cite{Swartout1993} listing `\textit{fidelty}', `\textit{understandability}', and `\textit{sufficiency}' as essential characteristics of an explanation. We contribute an instantiation of these desiderata in the context of counterfactual explanations (cf.\ Section \ref{section:desiderata} and Figure \ref{fig:intro}, right) as well as novel mechanisms to fulfill these desiderata. We combine these mechanisms into a novel algorithm called `smooth counterfactual explorer' (SCE), which we present in Section \ref{section:sce}. The effectiveness of our approach is demonstrated in Sections \ref{section:setup} and \ref{section:results} through evaluations on synthetic datasets with access to latent variables and real-world datasets. The actionability of SCE is further demonstrated through in-the-loop experiments, where we show that SCE can be used to effectively repair an ML model by reducing its reliance on spurious correlations and thereby improving its overall accuracy.

\section{Related Work}

\paragraph{Visual Counterfactual Explainers}
A variety of approaches have been proposed for producing counterfactuals of image classifiers, each of which aiming at overcoming various challenges with the optimization problem itself and with the quality of the resulting counterfactuals. Diffeomorphic Counterfactuals (DiffeoCF)~\cite{ dombrowski2022diffeomorphic} and DiVE~\cite{rodriguez2021beyond} aim to produce counterfactuals that remain on the data manifold by using a generative model, specifically an invertible latent space model.
The Adversarial Visual Counterfactual Explanations (ACE)~\cite{jeanneret2023adversarial} method uses gradients filtered through the noising/denoising process of a diffusion model~\cite{ho2020denoising} to avoid moving in adversarial directions.
It also uses a combination of L1 and L2 regularization between factual and counterfactual and RePaint~\cite{lugmayr2022repaint} as post-processing to keep pixel-wise changes to a minimum.
Unlike ACE, which uses one noise level and re-encodes the current counterfactual at each iteration, DVCEs~\cite{augustin2022diffusion, augustin2024dig}, DiME~\cite{jeanneret2022diffusion}, and FastDiME~\cite{weng2025fast} encode only once in the beginning to a latent state that is then updated based on classifier guidance~\cite{dhariwal2021diffusion}.
TIME~\cite{jeanneret2024text}, GCD~\cite{sobieski2024global}, and DiffEx~\cite{kazimi2024explaining} are also diffusion-based visual counterfactuals, but they are not based on the gradients of the classifier and are weaker on the metrics.
Compared to existing VCEs, our proposed approach combines a broader set of counterfactual generation mechanisms, with the aim of fulfilling more holistic desiderata.

\paragraph{Evaluating Explanations} A major challenge in designing explanation techniques is the issue of evaluation.
Ground-truth explanations are rarely available, and the notion of a good explanation depends on the use case.
The question of evaluation has been central, with early foundations including \cite{shapley:book1952} and \cite{Swartout1993}, which proposes practical desiderata for explanation techniques.
While the evaluation of explanations has been extensively covered for attribution methods (e.g.\ \cite{DBLP:journals/tnn/SamekBMLM17,DBLP:conf/nips/AdebayoGMGHK18,DBLP:conf/cvpr/BinderWLMMS23,DBLP:journals/csur/NautaTPNPSSKS23,DBLP:journals/jmlr/HedstromWKBMSLH23,DBLP:journals/tmlr/BluecherVS24}), there have been comparatively fewer such studies in the context of visual counterfactual explanations.
Guidotti et al.~\cite{guidotti2024counterfactual} proposed plausibility for linear models, which is similar to our concept of fidelity.
Mothilal et al.~\cite{mothilal2020explaining} proposed validity (a specific aspect of what we call fidelity), diversity, and sparsity in the context of linear counterfactuals on tabular data.
\cite{van2021interpretable} considers both sparsity and interoperability for visual counterfactuals.
Two measures (IM1 and IM2) of closeness to the data manifold were proposed for the latter purpose.
DiffeoCF~\cite{dombrowski2022diffeomorphic} focuses on IM1.
On CelebA, DiVE~\cite{rodriguez2021beyond} proposed to measure sparsity by counting the mean number of attributes changed (MNAC) in a counterfactual.
DiME~\cite{jeanneret2022diffusion} introduced measures of image quality based on the FID and/or sFID scores of the counterfactuals and a wide range of generic or domain-informed minimality metrics that DVCEs, ACE, FastDiME, TIME, and GCD also use.
DiME is the only one that measures diversity by measuring the LPIPS between multiple counterfactuals for the same factual.
In summary, while similar metrics to the one suggested by us have been proposed in the context of tabular data, previous work on visual counterfactual evaluation has predominantly focused on \textit{minimality} and adherence to the data manifold/\textit{image quality}, with the latter instantiated in an ad hoc fashion without deriving from formal desiderata.
\section{Desiderata of Counterfactuals}
\label{section:desiderata}

With the aim to enhance the usefulness of visual counterfactual explainers (VCEs), we propose to take as a starting point the holistic explanation desiderata formulated by Swartout and Moore (1993) \cite{Swartout1993}. These desiderata, originally proposed in the context of explaining expert systems, are `fidelity', `understandability', `sufficiency', `low construction overhead', and `[runtime] efficiency'. In the following, we contribute an instantiation of these desiderata, specifically the first three desiderata, to counterfactual explanations.

\subsection{Fidelity}

For an explanation to be useful, it should faithfully describe what the model does. As noted in \cite{Swartout1993}, an incorrect or misleading explanation is worse than no explanation at all. Translating the desideratum to counterfactual explanations, one requires that (1) the transformation of the data $x$ into its counterfactual $\widetilde{x}$ is plausible in practice, and that (2) the classifier $f$ responds strongly to this transformation and flips the classification. Most state-of-the-art VCEs, such as  DiffeoCF \cite{dombrowski2022diffeomorphic}, DiVE \cite{rodriguez2021beyond}, ACE \cite{jeanneret2023adversarial}, DiME \cite{jeanneret2022diffusion}, and FastDiME \cite{weng2025fast}, ensure the first aspect by limiting the search for $\widetilde{x}$ to the data manifold $\mathcal{M}$ through a \emph{generative model}. However, they do not address whether the transformation $x \mapsto \widetilde{x}$ is associated with a robust model response that allows crossing the decision boundary. Selecting any on-manifold transformation that flips the classifier (e.g.\ a minimal one) risks producing an adversarial example (cf.\ \cite{szegedy2016rethinking,DBLP:conf/cvpr/Stutz0S19}), as the counterfactual search may leverage spurious local variations in the classification function $f$ that are not representative of how $f$ varies more globally. Spurious local variations are commonplace in image-based classifiers (e.g.\ \cite{balduzzi2017,DBLP:journals/corr/SmilkovTKVW17,DBLP:journals/pieee/SamekMLAM21,DBLP:journals/corr/SzegedyZSBEGF13}), and also occur along the data manifold $\mathcal{M}$ \cite{DBLP:conf/cvpr/Stutz0S19}. Most existing counterfactual methods such as DiffeoCF, DiVE, ACE, DiME, and FastDiME lack a mechanism to address potential spurious variations on the manifold.

\subsection{Understandability}

Understandability refers to whether an explanation is understandable to its intended recipient, who is usually a human. Explanations should be presented at an appropriate level of abstraction and be concise enough to be quickly assimilated. In the context of counterfactual explanations, understandability can be characterized by the \textit{sparseness} of the transformation $x \mapsto \widetilde{x}$, which allows only a few features to change between $x$ and $\widetilde{x}$ in an interpretable latent space.
Mechanisms for sparseness can be found in several state-of-the-art VCEs, such as in ACE \cite{jeanneret2023adversarial} where sparseness is enforced trough a combination of $\ell_1$ losses between the factual and the counterfactual, or the so-called RePaint function~\cite{lugmayr2022repaint} which resets specific image regions of $\widetilde{x}$ the original pixel values found in $x$. These sparsity mechanisms contribute to make the counterfactual explanation more concise, thereby increasing its \textit{understandability}.

\subsection{Sufficiency}

In the context of explaining an ML model, sufficiency can be interpreted as an explanation's ability to provide enough information to make it actionable. In the context of counterfactuals, the sufficiency of an explanation can be promoted by generating for each instance $x$ a \textit{diverse} set of counterfactuals, highlighting the multiple, potentially interdependent factors that influence the decision function. This detailed account of the potentially many local effects is crucial for explanation-based model improvement in order to address the potentially multiple model flaws and build in the desired invariances. Diversification mechanisms can be found in the context of linear models of tabular data, with \cite{russell2019efficient} proposing to actively prevent the same features from being updated twice in two different counterfactual generation steps. However, such mechanisms to induce diversity are largely missing in existing visual counterfactual explainers such as DiffeoCF, ACE and DiME, except for the possibility to rerun counterfactual generation multiple times with different seeds.

\section{Smooth Counterfactual Explorer (SCE)}
\label{section:sce}

\begin{figure*}[t!]
\centering
\makebox[\textwidth][c]{\includegraphics[width=.95\textwidth]{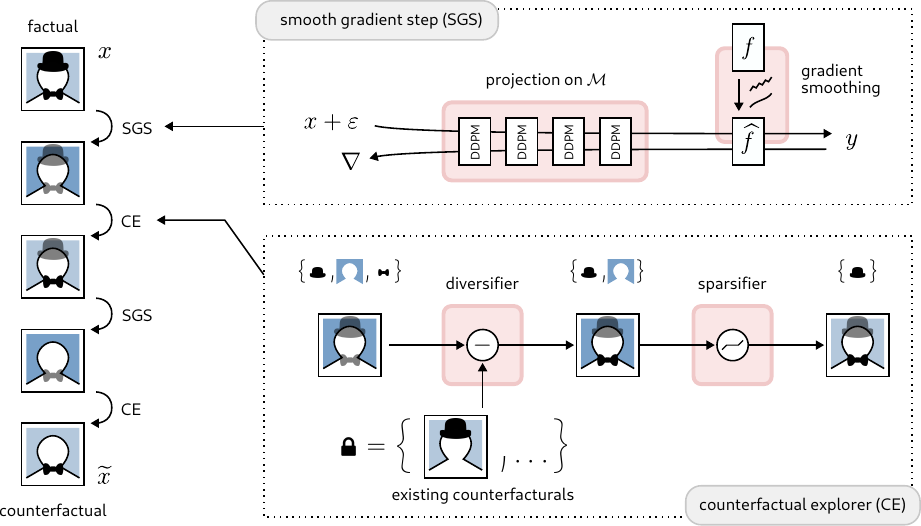}}
\caption{
Illustration of our proposed Smooth Counterfactual Explorer (SCE). It includes a diffusion model (DDPM) to project the data and gradients on manifold $\mathcal{M}$, a distillation of the original model into a surrogate model with smoother gradients, a sparsification of the transformation $x \mapsto \widetilde{x}$, and a lock-based diversifier forcing the currently generated counterfactual to differ from those generated previously. In the given example, we assume that the model $f$ responds to three features (`hat', `bowtie' and `background'). The bowtie is prevented from being removed in the current counterfactual because it was removed in the previous one. Of the remaining possible removals (hat and background), only the hat is removed in order to implement counterfactual sparsity.
}
\label{fig:diagram-sce}
\end{figure*}

We propose a novel counterfactual explainer that exhaustively addresses the desiderata of Section \ref{section:desiderata} through the incorporation of mechanisms found in existing counterfactual generation methods as well as novel mechanisms such as \textit{smooth distilled surrogates} and \textit{lock-based diversifiers}, which we introduce below. Our SCE approach is illustrated in Fig.\ \ref{fig:diagram-sce}. The detailed sequence of computations that SCE performs is given in Algorithm \ref{alg:pdc-simplified}. Table \ref{tab:mechanisms} shows how our SCE approach differs from existing methods in terms of implemented mechanisms and how these mechanisms help fulfill the desiderata.

\paragraph{Inducing Fidelity} Similar to methods such as DiffeoCF \cite{dombrowski2022diffeomorphic}, ACE \cite{jeanneret2023adversarial}, and DiME \cite{jeanneret2022diffusion}, we ensure a focus on plausible data transformation $x \mapsto \widetilde{x}$ by performing counterfactual search on the data manifold $\mathcal{M}$. Specifically, we pass the data through a denoising diffusion probabilistic model (DDPM)~\cite{ho2020denoising} that iteratively converts a noisy input into a denoised version, i.e.\ $\text{DDPM}(x) = (d \circ \dots \circ d)(x+\varepsilon)$ where $d$ is the decoding function. Propagating the gradient through the decoding stack suppresses part of the gradient $\nabla f$ that does not align with $\mathcal{M}$. To address the additional (and so far unsolved) problem of on-manifold spurious function variations, we propose conceptually to smooth the classifier's gradient field $\nabla f$, i.e.\ $\Vec{g} = (\nabla f) \ast k$ where $k$ is the smoothing kernel. To make this computationally feasible, we perform smoothing indirectly, by distilling $f$ into a surrogate model $\widehat f$ which we train to match $f$ under data augmentation (an adversary \cite{mkadry2017towards}, MixUp~\cite{zhang2017mixup} and label smoothing~\cite{szegedy2016rethinking}). To facilitate smoothing, we equip the student model with tailored nonlinearities combining LeakyReLU and Softplus. Performing counterfactual search on the gradient of the smoothed function $\widehat{f}$ allows SCE to avoid getting trapped in on-manifold local minima and to emphasize global variations of the classifier over local ones.

\begin{table*}[b!] \small \centering
\caption{Comparison of existing counterfactual methods and our SCE approach in terms of their mechanisms for fulfilling explanation desiderata. $\mathcal{M}$ denotes a manifold projection mechanism (e.g.\ a generative model), $\ell_1$ and \faPencilSquareO{} denote the addition of $\|x-\widetilde{x}\|_1$ as a penalty term and the RePaint function respectively, $\widehat{f}$ denotes our proposed function smoothing, and \faLock{} denotes our proposed lock-based diversification mechanism.}
\label{tab:mechanisms}
\medskip
\makebox[\textwidth][c]{
\begin{tabular}{lccccc}\toprule
  Desiderata & DiffeoCF \cite{dombrowski2022diffeomorphic} & DIME \cite{jeanneret2022diffusion} & ACE \cite{jeanneret2023adversarial} & FastDIME \cite{weng2025fast} & SCE \textbf{(ours)}\\\midrule
Fidelity & $ \mathcal{M}$ & $ \mathcal{M}$ & $\mathcal{M}$ & $\mathcal{M}$ & $\mathcal{M}+\widehat{f}$\\[.5mm]
Understandability (sparsity) & - & - & $\ell_1 +\text{\faPencilSquareO}$ & $\ell_1 +\text{\faPencilSquareO}$ & $\ell_1 +\text{\faPencilSquareO}$\\[.5mm]
Sufficiency (diversity) & - & - & - & - & \faLock{}\\
\bottomrule
\end{tabular}
}
\end{table*}

\paragraph{Inducing Understandability} Similar to ACE and FastDiME, we induce sparsity in SCE using the RePaint function and an $\ell_1$ penalty between the original and the counterfactual. Unlike ACE, which applies the RePaint function only after the counterfactual search ends, we apply it at each step of the counterfactual generation procedure. The proposed repeated application of the RePaint function in SCE enables us to perform the whole counterfactual search in the space of sparse counterfactuals, and to avoid that the last repainting step introduces sparsity at the expense of other desirable properties of the counterfactual $\widetilde{x}$.

\paragraph{Inducing Sufficiency} Our goal is to surpass the implicit diversification mechanism of ACE, DiME, and FastDiME, which rely on random factors in the optimization procedure, by actively pulling apart the different generated counterfactuals. We propose a novel \textit{lock-based diversifier}, where counterfactual updates are only allowed to follow directions that differ from previously generated counterfactuals. This locking mechanism is enforced in the gradient descent step in both latent and pixel spaces. (In pixel space, this is achieved through the RePaint function.) As a final step, the sequence of generated counterfactuals is clustered and ranked based on counterfactuals obtained from the entire dataset. Specifically, reranking involves clustering the counterfactual direction vectors in latent space using k-means, with a cosine similarity variant. The clusters are then ranked according to the average of direction vectors within each cluster (the cluster with the largest vector is ranked first). Finally, the counterfactuals are ranked according to the rank of the clusters to which they belong.

\begin{algorithm}[t!]
    \caption{Smooth Counterfactual Explorer (SCE). It takes as input a classifier $f$, a diffusion model $d$ and a data point $x$ assumed to be positively classified ($f(x)>0)$. It outputs a sequence of counterfactuals $(\widetilde{x}_1,\dots,\widetilde{x}_K)$ that explain the prediction $f(x)$.}
    \label{alg:pdc-simplified}
    \begin{algorithmic}
    \STATE $\widehat{f} \gets \text{smooth}(f)$
    \FOR{$k = 1 \dots K$}
        \STATE $z \gets x$
        \REPEAT
            \STATE $\text{\scriptsize\faLock} = \{\widetilde{x}_1,\dots,\widetilde{x}_{k-1}\}$ 
            \STATE $z = z - \gamma \cdot \nabla(\widehat{f} \circ d \circ \dots \circ d)(z+\varepsilon)$
            \STATE $z \gets \text{exclude}(z,x, \text{\scriptsize\faLock})$
            \STATE $z \gets \text{sparsify}(z,x)$
        \UNTIL{$f(z) < \tau$}
        \STATE $\widetilde{x}_k \gets z$
    \ENDFOR
    \STATE {\bfseries return} \,$\text{cluster\_and\_rank}(\widetilde{x}_1,\dots,\widetilde{x}_K)$
    \end{algorithmic}
\end{algorithm}

\paragraph{Computational Aspects}

To generate a set to up to $K$ counterfactuals associated to one data point $x$ and denoting by $T$ the number of gradient descent step per counterfactual, SCE requires up to $K \times T$ forward and backward passes through the generative model and the same number of forward passes for the RePaint in the sparsifier. 
Typically, the value of $K$ in SCE is low (between $1$ and $2$ in our experiments) as only few distinct counterfactual can be generated for a given factual $x$. Per counterfactual, SCE has the same complexity $T$ as ACE. However, compared to ACE, the amount of computations varies in the following way: (1) After every iteration, SCE checks whether the output of the classifier $f$ has flipped and stops early if it has been achieved. At the same time, ACE typically performs gradient descent for a fixed number of iterations. (2) Due to the smoother gradients from our gradient smoothing, the counterfactual search of SCE converges faster than ACE. (3) Compared to ACE, SCE includes the initial phase of building the distilled model $\widehat{f}$ before being able to compute the counterfactuals.

\section{Experimental Setup}
\label{section:setup}

In this section, we first describe the collection of datasets and tasks on which we conduct our experiments. Then, we propose a set of metrics to measure whether the tested counterfactual explainers fulfill the desiderata of Section \ref{section:desiderata}.
Next, we perform benchmark experiments comparing our SCE method with a set existing VCEs, specifically, ACE, DiME, and FastDiME, as well as an ablation study that tests the necessity of each component of SCE. Our quantitative results are complemented with qualitative results where we provide visual interpretation of the advantage of our SCE approach.

\subsection{Datasets and Models}
\label{lab:datasets}

We select various models and classification tasks, ranging from natural to purely synthetic. Our selection aims to verify the robustness of SCE in a broad range of settings, and to provide a variety of ground truths for evaluation purposes. For all our experiments, we used as a generative model the DDPM implementation of Dhariwal et al.~\cite{dhariwal2021diffusion}.

First, we consider a ResNet18~\cite{he2016deep} with weights from torchvision fine-tuned on the CelebA dataset \cite{liu2015faceattributes}. The CelebA dataset is a collection of face images, each of which is categorized by 40 different attributes. We fine-tune the torchvision weights on the CelebA `smiling' attribute, and use the resulting model for the task of generating smiling to non-smiling counterfactuals and vice versa. We repeat the experiment for the `blond' attribute, which differs from the smiling attribute by its significantly larger pixel footprint. One advantage of the CelebA dataset is that ground-truth latent factors can be generated. In our experiments, we use the 40 logits of a CelebA-pretrained DenseNet\footnote{https://github.com/ServiceNow/beyond-trivial-explanations} for that purpose.
Additionally, we run experiments on an augmented version of CelebA, where we add a `copyright tag'-like visual artifact to the bottom right corner, in a way that it spuriously correlates with the `smiling' attribute (see Supplementary Figure 1). This semi-synthetic dataset provides ground-truth for our evaluation as the location of the injected copyright tag is known. Specifically, we have access to segmentation masks that separate the foreground from the copyright tag location. We also repeat the same experiment with a different model: a vision transformer~\cite{dosovitskiy2020image}.

Next, we consider a ResNet18 with weights from torchvision and finetuned on a purely synthetic dataset called `\textit{Square}'. The Square dataset (See Supplementary Figure 2) consists of a simple 4-dimensional manifold embedded in image space. The four latent dimensions are the intensity of the foreground square, the intensity of the background, and the x- and y-position of the foreground square. A spurious correlation is introduced between the (relevant) foreground square's intensity and the (irrelevant) background square's intensity. This spurious correlation, combined with the background's higher saliency (high pixel footprint), causes the model to learn a Clever Hans strategy based on the background (cf.\ \cite{hermann2023foundations} for a related study). This dataset has direct access to the latent features, namely the square's position and the background/foreground intensities, enabling further scrutiny into the correctness of the counterfactuals.

Finally, we consider the ResNet18 with weights from torchvision,  this time fine-tuned on a subset of the Camelyon17 dataset. This subset introduces a spurious correlation between the histopathological patch type (benign or malignant) and the hospital from which the patch originates. This simulates a situation in which all malignant samples come from one hospital and all benign samples come from another. We access the two-dimensional semantic latent space by training an oracle model on the full dataset, which has no such correlation, to predict both the patch type and the hospital. We then use the oracle's logits as the latent representation.

\subsection{Metrics for Testing the Desiderata}
\label{section:metrics}

We now introduce several metrics for testing the fulfillment of the desiderata discussed in Section \ref{section:desiderata}.
As a starting point, we need to test whether a given method is capable of generating counterfactuals at all. We quantify this by the `flip rate' 
$\mathrm{FR}=N_\text{flipped}/N_\text{total}$
where $N_\text{flipped}$ is the number of counterfactuals that flipped the decision of the explained predictor and $N_\text{total}$ is the number of data points where a counterfactual was attempted.
The following metrics focus on examples that are successfully flipped.
First, we want to verify that the produced counterfactuals implement \textit{fidelity}, i.e.\ follow the main variations of the prediction function.
To verify specifically that counterfactuals robustly cross the decision boundary, we test our predictions against a new model distilled from the original model, where we use a different seed for the weight initialization. This allows us to verify that generated counterfactuals are not fooled by weight-specific adversarial attacks and can be quantified by the non-adversarial rate (NA):
\begin{align}
\text{Fidelity:NA}= N_{\text{flipped\_true}} / N_{\text{flipped}}
\label{eq:fidelity-na}
\end{align}
The latter counts how often the decision boundary is also crossed in the distilled model. The higher the NA score, the more faithful the counterfactual. We consider a second metric of fidelity that verifies that the counterfactual explainer focuses on primary variations. First, we identify the dominant feature (e.g.\ with large pixel footprint) and compute the rate $R_\mathrm{true}$ at which it dominates the competing feature (e.g.\ with small pixel footprint). Then, we compute the rate $R_\mathrm{actual}$ at which the dominant feature changes in the counterfactual. Our proposed `dominant rate' metric is then defined as the ratio of these two quantities:
\begin{align}
    \text{Fidelity:DR}=R_\text{actual}/R_\text{true}
    \label{eq:fidelity-dr}
\end{align}
As a test for \textit{understandability}, we define $e$ an encoding function mapping the data to some meaningful latent space (e.g.\ the 40-dimensional space of attributes in CelebA) and represent the difference between factual and counterfactual as $\Delta = e(x)-e(\widetilde{x})$. We then quantify understandability as the sparsity of the difference vector $\Delta$:
\begin{align}
\text{Sparsity}=1-\mathrm{avg}(|\Delta|)/{\max}(|\Delta|)
\label{eq:sparsity}
\end{align}
where $|\cdot|$ applies element-wise, `${\max}$' takes the maximum over this vector, and `$\mathrm{avg}$' calculates the average over this vector, excluding the maximum. A score of $0$ indicates a maximally nonsparse (i.e.\ constant) vector, and a score of $1$ indicates a maximally sparse vector. 
The sparsity score is then averaged over all generated counterfactual instances.
We then test whether the counterfactual techniques achieve \textit{sufficiency}, i.e.\ produce a sufficiently rich explanation by measuring the diversity of counterfactuals they produce for each data point.
As for sparsity, we measure diversity on the dataset-specific latent differences. Specifically, we generate two counterfactuals for each original sample and compute:
\begin{align}
\text{Diversity}=1-\mathrm{CosSim}(\Delta_1,\Delta_2)
\label{eq:diversity}
\end{align}
where $\Delta_1$ and $\Delta_2$ denote the differences (in latent space) between the factual and the two generated counterfactuals.

\subsection{In-the-Loop Gain}
\label{section:gain}

To evaluate the actionability of the different VCEs, we integrate them in the recently proposed CFKD~\cite{bender2023towards} evaluation framework. CFKD simulates an environment containing a VCE agent and a user agent. The simulated user agent is given oracle knowledge, receives the generated counterfactuals as input and outputs data consolidation proposals on which the function $f$ can be retrained. We apply this evaluation to all settings where the data has a spurious correlation (see Section~\ref{lab:datasets}) and the model an associated Clever Hans strategy. The objective we set for the VCE is to enable the model to get rid of its Clever Hans strategy through meaningful data consolidation steps. The performance metric in the CFKD environment is the unpoisoned test accuracy (i.e.\ where the data is stripped from its artificial spurious correlation). We compare the unpoisoned test accuracy before and after applying CFKD and calculate the \textit{Gain} as how much less likely the model is to make a classification error after running CFKD:
$$\mathrm{Gain} = \frac{\mathrm{Acc}_\mathrm{after} - \mathrm{Acc}_\mathrm{before}}{\mathrm{Err}_\mathrm{before}}$$
Positive Gain indicates that the explanations contain useful information for identifying and fixing errors in a model. Our in-the-loop gain evaluation can also be viewed as a simulation of a human study, with the difference that the user is modeled as an oracle and the study is fully reproducible. Furthermore, measuring performance gain rather than relying on subjective human feedback prevents logical fallacies in the study design, ensuring that explanations useful for improving the model are ranked higher than those that the user merely expects.

\section{Results}
\label{section:results}

\begin{table*}[t!]
\centering \footnotesize
\renewcommand{\arraystretch}{1.1}
\setlength{\tabcolsep}{6pt}
\begin{tabular}{llrcccccc}
\toprule
& & & & \multicolumn{4}{c}{\cellcolor{CadetBlue!20}\textbf{desiderata}}\\
& & & & \multicolumn{2}{c}{\cellcolor{CadetBlue!20}{fidelity}} & \cellcolor{CadetBlue!20}{underst.} & \cellcolor{CadetBlue!20}{sufficiency} & \\
Dataset / Model & Method & & ~~~\textit{FR}~~~ & \cellcolor{CadetBlue!20}(NA) & \cellcolor{CadetBlue!20}(DR) & \cellcolor{CadetBlue!20}(sparsity) & \cellcolor{CadetBlue!20}(diversity) & gain  \\\midrule
\multirow{4}{*}{CelebA-Smile / ResNet-18} & ACE                  & & \textit{99.0} & \textbf{98.0} & - & \textbf{78.9} & 19.2 & - \\
 & DiME                                                     & & \textit{100.0} & 91.0 & - & 73.0 & \underline{42.4} & - \\
 & FastDiME                                                 & & \textit{100.0} & 78.0 & - & 70.7 & 38.8 & - \\
\rowcolor{gray!10}\cellcolor{white} & SCE \textbf{(ours)}                                               & & \textit{81.0} & \underline{92.6} & - & \underline{76.3} & \textbf{72.9} & -  \\\midrule
\multirow{4}{*}{CelebA-Blond / ResNet-18} & ACE               & & \textit{51.0} & \underline{80.4} & - & \underline{75.8} & 46.6 & - \\
 & DiME                                                     & & \textit{92.0} & 64.1 & - & 72.3 & \underline{51.9} & - \\
 & FastDiME                                                 & & \textit{75.0} & 52.3 & - & 72.2 & 39.2 & - \\
\rowcolor{gray!10}\cellcolor{white} & SCE \textbf{(ours)}                                               & & \textit{91.0} & \textbf{97.8} & - & \textbf{79.3} & \textbf{75.5} & - \\\midrule
\multirow{4}{*}{CelebA-Smile\,+\,\copyright{} / ResNet-18} & ACE  & & \textit{34.0} & \underline{55.9} & 70.0 & \underline{29.3} & 1.8 & \underline{37.8} \\
 & DiME                                                     & & \textit{88.5} & 12.4 & 74.7 & 4.3 & 0.0 & 3.0 \\
 & FastDiME                                                 & & \textit{57.5} & 14.8 & \textbf{86.6} & 7.85 & 0.1 & 8.25 \\
\rowcolor{gray!10}\cellcolor{white} & SCE \textbf{(ours)}                                               & & \textit{94.0} & \textbf{100.0} & \underline{81.7} & \textbf{72.5} & \textbf{30.9} & \textbf{69.6} \\\midrule
\multirow{4}{*}{CelebA-Smile\,+\,\copyright{} / ViT-16B} & ACE    & & \textit{59.0} & \underline{54.6} & \textbf{100.0} & \underline{64.8} & \underline{0.5} & \underline{16.0} \\
 & DiME                                                     & & \textit{90.0} & 15.6 & 87.5 & 5.2 & 0.0 & -0.4 \\
 & FastDiME                                                 & & \textit{75.0} & 10.7 & 91.7 & 7.9 & 0.1 & 0.2 \\
\rowcolor{gray!10}\cellcolor{white}  & SCE \textbf{(ours)}                                               & & \textit{66.5} & \textbf{63.9} & \textbf{100.0} & \textbf{95.1} & \textbf{22.7} & \textbf{30.9} \\\midrule
\multirow{4}{*}{Square / ResNet-18} & ACE                & & \textit{3.40} & \textit{100.0} & \textit{36.7} & \textit{99.5} & \textit{0.0} & \textit{0.0}\\
 & DiME                                                     & & \textit{31.9} & 28.4 & 0.0 & \underline{92.0} & 0.4 & \underline{9.3}\\
 & FastDiME                                                 & & \textit{24.0} & \underline{36.7} & \underline{7.38} & 87.9 & \underline{1.6} & -3.3 \\
\rowcolor{gray!10}\cellcolor{white} & SCE \textbf{(ours)}                                               & & \textit{100.0} & \textbf{96.1} & \textbf{103.9} & \textbf{95.4} & \textbf{95.5} & \textbf{93.4} \\\midrule
\multirow{4}{*}{Camelyon17 / ResNet-18} & ACE               & & \textit{56.5} & 29.2 & 12.8 & 55.0 & \underline{18.6} & 7.0 \\
 & DiME                                                     & & \textit{93.0} & 44.6 & 2.1 & 45.9 & 8.3 & -3.8 \\
 & FastDiME                                                 & & \textit{77.5} & \underline{60.0} & \underline{64.5} & \underline{55.2} & 8.2 & \underline{16.3} \\
\rowcolor{gray!10}\cellcolor{white}  & SCE \textbf{(ours)}                                               & & \textit{53.0} & \textbf{100.0} & \textbf{64.8} & \textbf{58.0} & \textbf{33.6} & \textbf{22.0} \\
\bottomrule
\end{tabular}
\caption{Desiderata-driven evaluation of our proposed SCE approach. We compare our approach to three competing approaches across six different dataset/architecture combinations. The four numerical columns in the middle quantify the extent to which each desiderata is fulfilled according to Eq.\ \eqref{eq:fidelity-na}.  Higher is better. Additionally, the leftmost numerical column provides nominal flip rates for indicative purposes.  The rightmost column shows the end-to-end metric `gain' which is described in Section \ref{section:gain}. Higher scores indicate more actionable explanations.}
\label{tab:quantiative_results}
\end{table*}

Using on the quality metrics above, we proceed with comparing four VCEs: ACE \cite{jeanneret2023adversarial}, DiME \cite{jeanneret2022diffusion}, FastDIME \cite{weng2025fast}, and the proposed Smooth Counterfactual Explorer (SCE). We first look at counterfactual quality in terms of the \textit{fidelity}, \textit{understandability} and \textit{sufficiency} desiderata, as quantified by Eqs.\ \eqref{eq:fidelity-na}--\eqref{eq:diversity}. Results are shown in the corresponding columns in Table \ref{tab:quantiative_results}. We also report the nominal flipping rates (column `FR') for reference. Examining the Fidelity:NA scores, we observe that our SCE approach performs robustly across all datasets and ranks first in 5 out of 6 considered datasets. This highlights the higher immunity of SCE to adversarial counterfactuals. ACE fares best on the CelebA-Smile but tends to not exhibit robust performance on other datasets. On the Fidelity:DR metric, we similarly observe the higher robustness of SCE at systematically at focusing on dominant features (with a large pixel footprint). In comparison, ACE, DiME, and FastDiME are less predictable, and on some datasets, tend to systematically turn to secondary features (with a small pixel footprint). In terms of \textit{understandability}, SCE performs on par with ACE, which can be explained by their common sparsity mechanism based on the RePaint function.  SCE and ACE sparsity capabilities also appear significantly more robust than DiME and FastDiME. Looking at explanation \textit{sufficiency}, SCE scores above competitors on the diversity metric by a wide margin, which can be attributed to our newly proposed lock-based diversifier and the absence of comparable mechanisms in other counterfactual explainers. 
Considering then our \textit{Gain} metric, SCE again outperforms the other VCEs by a wide margin on all datasets. We see this as a natural consequence of fulfilling the various explanation desiderata, and this also demonstrates that our desiderata-driven approach is particularly effective in producing explanations that are actionable.
Lastly, the performance of SCE appears fairly stable w.r.t.\ the choice of model to explain, as shown by similar qualitative results when replacing ResNet-18 by a ViT-16B model.

\begin{table*}[t!]
\centering \footnotesize
\renewcommand{\arraystretch}{1.1}
\setlength{\tabcolsep}{8pt}
\begin{tabular}{llcccccc}
\toprule
& & & \multicolumn{4}{c}{\cellcolor{CadetBlue!20}\textbf{desiderata}}\\
& & & \multicolumn{2}{c}{\cellcolor{CadetBlue!20}{fidelity}} & \cellcolor{CadetBlue!20}{underst.} & \cellcolor{CadetBlue!20}{sufficiency} & \\
Mechanisms & & ~~~\textit{FR}~~~ & \cellcolor{CadetBlue!20}(NA) & \cellcolor{CadetBlue!20}(DR) & \cellcolor{CadetBlue!20}(sparsity) & \cellcolor{CadetBlue!20}(diversity) & gain \\\midrule
\phantom{$\mathcal{M}$,} $\ell_1 +\text{\faPencilSquareO}$, $\widehat{f}$, \faLock{}         & (no manifold proj.) & \textit{48.5} & 41.4 & 55.1 & 88.8 & \underline{74.8} & 32.8 \\
$\mathcal{M}$, \phantom{$\ell_1 +\text{\faPencilSquareO}$,} $\widehat{f}$, \faLock{}         & (no sparsity) & \textit{100.0} & \underline{93.6} & 25.8 & 61.7 & 6.5 & 68.7 \\
$\mathcal{M}$, $\ell_1 +\text{\faPencilSquareO}$, \phantom{$\widehat{f}$,} \faLock{}        & (no smoothness) & \textit{2.0} & \textit{50.0} & \textit{0.0} &\textit{ 100.0} & \textit{37.0} & 0.0 \\
$\mathcal{M}$, $\ell_1 +\text{\faPencilSquareO}$, $\widehat{f}$, \phantom{\faLock{}}       & (no diversity) & \textit{98.0} & 89.0 & \underline{90.5} & \underline{94.0} & 7.4 & \underline{89.9} \\
\rowcolor{gray!10}
$\mathcal{M}$, $\ell_1 +\text{\faPencilSquareO}$, $\widehat{f}$, \faLock{}                   & \textbf{(ours)}  & \textit{100.0} & \textbf{96.1} & \textbf{103.9} &  \textbf{95.4} & \textbf{95.5} & \textbf{93.4} \\
\bottomrule
\end{tabular}
\caption{Ablation study on the `square' dataset. The first four rows show the effect of deactivating each SCE mechanism individually. The last row corresponds to our original proposed SCE approach. The evaluation metrics are the same as in Table \ref{tab:quantiative_results}. The best results are in bold. The setting without smoothing $\widehat{f}$ is excluded from our ranking due to its extremely low flipping rate (FR).
}
\label{tab:ablation_study}
\end{table*}

\paragraph{Ablation Study} To verify the combined importance of the four mechanisms contained in SCE, namely the gradient filtering through the DDPM generative model (projection on the data manifold $\mathcal{M}$), the sparsifier  ($\ell_1 +\text{\faPencilSquareO}$), the smooth distilled surrogate ($\widehat{f}$) and the lock-based diversification mechanism (\faLock{}), we perform an ablation study, where each of these mechanisms is removed individually. The results are shown in Table \ref{tab:ablation_study} for the Square dataset. We observe that combining these four mechanisms is necessarily to fulfill the explanation desiderata. In particular, removing the data manifold projection mechanism greatly exposes the counterfactual generator to adversarial counterfactuals, as shown by a significant drop in the NA fidelity metric, and also reduction in the overall gain metric. Deactivating the sparsifier leads to an expected decrease in the sparsity score, a sharp reduction in diversity scores, and a significant reduction in the gain metric. Deactivating the smoothing mechanism breaks the counterfactual explainer, with the latter now failing to generate counterfactuals (as shown by a FR close to 2\%). Finally, as expected, deactivating the diversifier (and replacing it with a basic randomized diversifier) incurs a sharp drop in diversity, and a noticeable, although less drastic, reduction in the gain metric.

\begin{figure*}[t!]
\centering
    \makebox[\textwidth][c]{\includegraphics[width=\textwidth]{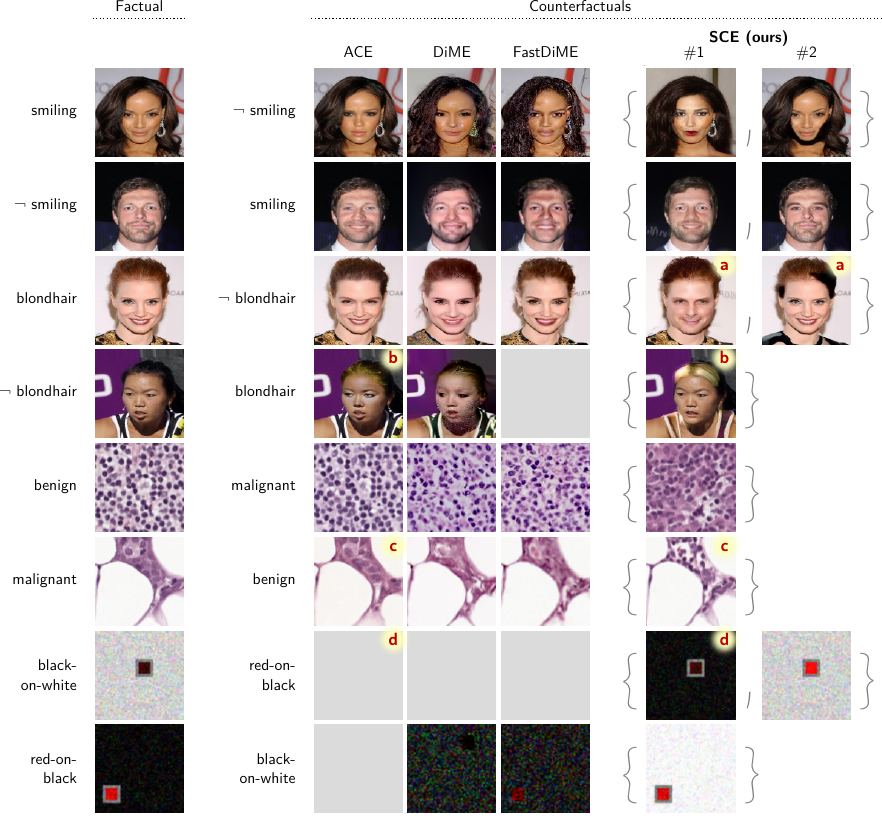}}
    \caption{Visualization of the counterfactuals generated by different methods. For a selection of tasks and classes, we display one correctly classified image alongside the counterfactuals produced by each method. For ACE, DiME, and FastDiME, we display one counterfactual generated with a given random seed. For SCE, we display as many counterfactuals as could be generated under the diversifier mechanism (here, one or two).
    As can be seen in (a), SCE is the only method that creates diverse counterfactuals, clearly highlighting different influential features.
    In (b) and (c), one can see that SCE emphasizes sparse changes instead of focusing on minimal, but not very interpretable changes.
    In (d), one can see that SCE and its gradient smoothing mechanism can better explore the input domain for true counterfactuals, whereas other methods get stuck on plateaus or local minima of the classifier $f$.
    }
    \label{fig:qualitative}
\end{figure*}

\paragraph{Qualitative Comparison} We present some qualitative results for the analyzed methods in Figure~\ref{fig:qualitative}. Here, we select two examples for each dataset (one from each class). To visualize the diversity capability of SCE, we generate as many counterfactuals as can be generated under the constraints imposed by SCE's diversifier mechanism, and these counterfactuals are ordered from left to right according to the result of the cluster-and-rank procedure. We observe that all methods work quite well for flipping the `smiling' attribute. However, the difficulty increases for the other four tasks as the distance between classes also increases. While ACE is often nominally successful at flipping counterfactuals, its lack of adversarial robustness leads it to generate instances that are not easily distinguishable from the original images, potentially misleading a human's understanding of the model. DiME and FastDiME yield results qualitatively similar to ACE. In contrast, SCE makes semantically more distinct changes to the factual, e.g., successfully changing the background of the squares dataset. The ability of SCE to produce a diverse set of counterfactuals, often revealing the multiple strategies contained in the classifier, is also highlighted. For example, on the smiling task, SCE reveals two different ways of reducing the smile, either by covering it with heavy makeup or by outlining the face with black color without touching the smile. Similarly, in the `blond hair' task, the `blond' attribute seems to be modifiable either by an actual color change or by tampering with the male/female features, thereby revealing a Clever Hans strategy \cite{lapuschkin2019unmasking}.
\section{Conclusion}

Counterfactuals are a promising approach to explain machine learning models, e.g.\ for image classification. In this paper, we analyzed  shortcomings of existing visual counterfactual explainers (VCEs) from the perspective of their explainability properties. Our experimental evaluation revealed significant heterogeneity in performance of VCEs across different models and datasets. Since counterfactual generation is essentially a nonlinear, nonconvex optimization problem, it inherits challenges such as escaping plateaus and finding global minima. These are challenges that other explanation methods, such as attribution \cite{DBLP:journals/kais/StrumbeljK14,bach2015pixel,sundararajan2017axiomatic}, typically do not face. Starting from the a well-established explanation desiderata \cite{Swartout1993} and proposing an instantiation to counterfactual explanations, we have contributed novel mechanisms for VCEs designed to improve their explanation capabilities. These new mechanisms were combined into a new algorithm, Smooth Counterfactual Explorer (SCE), and we demonstrated the benefits of our approach in terms of explanation quality and actionability, both qualitatively and quantitatively.

While our SCE approach improves the quality of generated counterfactuals, searching for them remains a challenging endeavor that requires optimizing a complex, nonlinear objective. Currently, this optimization is achieved through a subtle combination of multiple mechanisms and the use of auxiliary models, such as generators and distilled surrogates. Given these persistent optimization challenges, a promising area of future research is the intersection of counterfactual generation and attribution-based explanation techniques. For example, attribution techniques like LRP \cite{bach2015pixel} associate each prediction of a classifier $f$ with a simple local surrogate of $f$. These simple local surrogates could, in turn, ease the counterfactual search. Conversely, generated counterfactuals could help resolve open questions in attribution-based explanations, such as the question of reference points (cf.~\cite{letzgus2022toward}). Counterfactuals could serve as candidate reference points from which more precise attribution-based explanations could be obtained.

\section*{Acknowledgements}
This work was in part supported by the Federal
German Ministry for Education and Research (BMBF) under Grants
BIFOLD24B, BIFOLD25B, 01IS18037A, 01IS18025A, and 01IS24087C. K.R.M.\ was partly supported by the Institute of Information \& Communications Technology Planning \& Evaluation (IITP) grants funded by the Korea government (MSIT) (No. 2019-0-00079, Artificial Intelligence Graduate School Program, Korea University and No. 2022-0-00984, Development of Artificial Intelligence Technology for Personalized Plug-and-Play Explanation and Verification of Explanation).  Moreover, it was supported by BASLEARN- TU Berlin/BASF Joint Laboratory, co-financed by TU Berlin and BASF SE. 
We would also like to thank Lorenz Linhardt and Jonas Dippel for the valuable discussions and helpful comments.

\bibliographystyle{elsarticle-num}
\bibliography{main}

\includepdf[pages=-,offset=0 0]{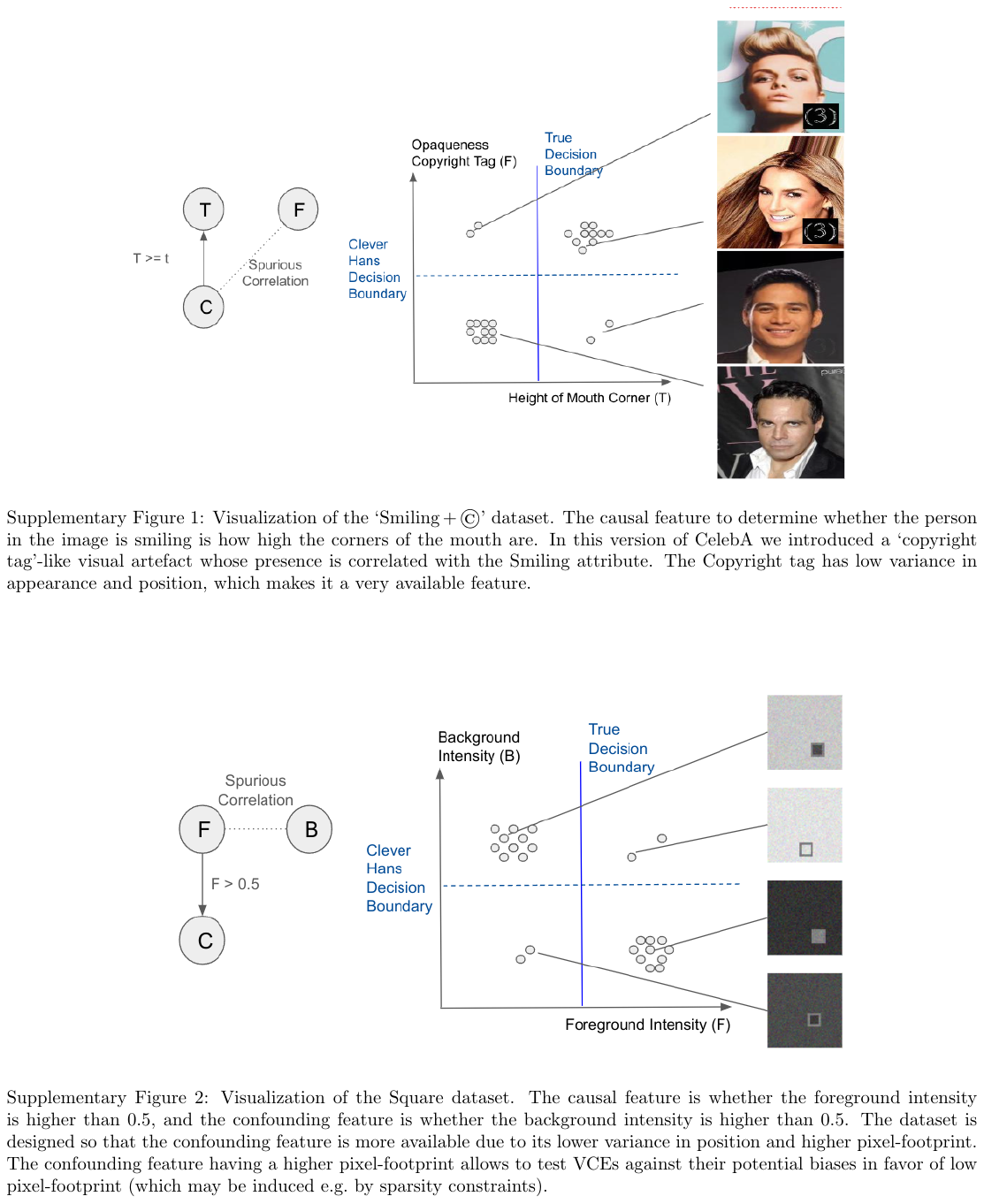}

\end{document}